\documentclass{article}

\usepackage{arxiv}
\usepackage{amssymb}
\usepackage{amsmath}
\usepackage{bm}
\usepackage{subfig}
\usepackage{wrapfig}
\usepackage{lipsum}     
\usepackage{float} 
\usepackage{graphicx}           
\usepackage{url} 
\usepackage{soul}
\usepackage{enumitem}
\usepackage{lineno}
\usepackage{diagbox}
\usepackage{color}
\usepackage{booktabs} 
\usepackage{xcolor}

\def\R{{\mathbb R}}

\def\curlyF{{\mathcal F}}

\def\curlyG{{\mathcal G}}

\def\curlyL{{\mathcal L}}

\def\curlyX{{\mathcal X}}
\def\curlyY{{\mathcal Y}}

\usepackage[ruled,vlined,linesnumbered,noresetcount]{algorithm2e}

\title{Scalable Uncertainty Quantification for Deep Operator Networks using Randomized Priors}

\author{
  Yibo Yang \\
  Department of Mechanical Engineering \\
  and Applied Mechanics\\
  University of Pennsylvania\\
  Philadelphia, PA 19104 \\
  \texttt{ybyang@seas.upenn.edu} \\
   \And
    Georgios Kissas \\
  Department of Mechanical Engineering \\
  and Applied Mechanics\\
  University of Pennsylvania\\
  Philadelphia, PA 19104 \\
  \texttt{gkissas@seas.upenn.edu } \\
   \And
  Paris Perdikaris \\
  Department of Mechanical Engineering \\
  and Applied Mechanics\\
  University of Pennsylvania\\
  Philadelphia, PA 19104 \\
  \texttt{pgp@seas.upenn.edu} \\
}

\begin{document}
\maketitle

\begin{abstract}
We present a simple and effective approach for posterior uncertainty quantification in deep operator networks (DeepONets); an emerging paradigm for supervised learning in function spaces. We adopt a frequentist approach based on randomized prior ensembles, and put forth an efficient vectorized implementation for fast parallel inference on accelerated hardware. Through a collection of representative examples in computational mechanics and climate modeling, we show that the merits of the proposed approach are fourfold. (1) It can provide more robust and accurate predictions when compared against deterministic DeepONets. (2) It shows great capability in providing reliable uncertainty estimates on scarce data-sets with multi-scale function pairs. (3) It can effectively detect out-of-distribution and adversarial examples. (4) It can seamlessly quantify uncertainty due to model bias, as well as noise corruption in the data. Finally, we provide an optimized JAX library called {\em UQDeepONet} that can accommodate large model architectures, large ensemble sizes, as well as large data-sets with excellent parallel performance on accelerated hardware, thereby enabling uncertainty quantification for DeepONets in realistic large-scale applications. All code and data accompanying this manuscript will be made available at \url{https://github.com/PredictiveIntelligenceLab/UQDeepONet}.

\end{abstract}


\section{Introduction}\label{sec:intro}

Deep learning and data-driven methods have found great success in scientific applications such as image classification \cite{krizhevsky2012imagenet}, language modeling \cite{hochreiter1997long}, reinforcement learning \cite{sutton2018reinforcement}, chemical discovery \cite{gomez2016automatic}, finance \cite{chaboud2014rise}, etc. In  recent years, the combination of artificial neural networks, statistical tools and domain knowledge in the modeling of physical systems has also received a lot of attention in sparse system identification \cite{brunton2016discovering}, solution of high-dimensional partial differential equations \cite{han2018solving}, and creating physics-informed data assimilation tools \cite{raissi2019physics}. 

A new direction in scientific deep learning aims at approximating maps between spaces of functions \cite{chen1995universal, kissas2022learning, lu2021learning, kovachki2021neural}, opening the path to building fast surrogate models that can be trained and queried in a continuous fashion at any resolution. One prominent approach for  supervised learning under this setting is the so-called deep operator network architecture (DeepONet) \cite{lu2021learning}, proposed by Lu \textit{et. al.} \cite{lu2021learning}. The DeepONet architecture constitutes an extension of the original architecture of Chen \textit{et. al.}, and has been successfully employed across different applications, including the prediction of boundary layers \cite{di2021deeponet}, bubble growth dynamics \cite{lin2021operator}, chemical reactions \cite{mao2021deepm}, and others. Another prominent approach for learning in function spaces is the family of Neural Operator methods proposed by Li \textit{et. al.} \cite{li2020neural, li2020fourier, li2018learning}, and further formalized mathematically by Kovachki \textit{et. al.} \cite{kovachki2021neural}. In this line of work, the authors propose an architecture that models the Green's function of  differential equations via functional compositions consisting of a linear transformation, an integral operator with a learnable parametric kernel, followed by a point-wise non-linearity. A more recent approach was proposed in Kissas \textit{et. al.} \cite{kissas2022learning} where the authors drew motivation from the success of the attention mechanism in deep learning \cite{bahdanau2014neural} to propose a novel kernel-coupled attention mechanism that is suitable for learning in function spaces. Their proposed model is constructed by taking the expectation of a feature vector extracted via a spectral encoder and a distribution built using the attention mechanism, with an additional coupling of the score functions for different query locations.

Despite the recent emergence of operator learning methods as a powerful tool across diverse applications, there are still open challenges that need to be addressed. First, initial results in \cite{kissas2022learning} suggest that the predictions of different operator learning methods provide distributions of errors which may contain a large number of outliers depending on the difficulty of the problem and the amount of labeled data used during training. Second, recent works \cite{wang2021improved, di2021deeponet} suggest that a bias might be present in a DeepONet model predictions, when trained on data-sets that include multi-scale target functions. Without carefully weighting the loss function, the operator learning model would over-fit to the high magnitude components of the target functions while poorly approximating others, due to imbalanced and vanishing gradients. Third, the majority of operator learning methods in the literature assume enough input/output function samples for training, which may not be always feasible in applications where the cost of data acquisition is high. For cases where only a small number of input/output function samples are available, the predictions might be poor and the prediction errors large. Fourth, when querying new input functions to approximate new solutions, out-of-distribution or adversarial samples might exist in the test data-set, inevitably leading to poor predictions. All these challenges suggest that using deterministic methods when learning from distributions of functions may lack reliability, thereby posing the need for methods that can account for uncertainty.

The field of uncertainty quantification deals with the presence of incomplete information in various stages of the scientific method. Sources of uncertainty can vary \cite{psaros2022uncertainty}, but most commonly include randomness introduced by partial observability or inherent randomness in the data-generating process, a limited amount of training labels, biased assumptions, or incomplete physical models, to list a few. There exist two general approaches in modeling uncertainty: the frequentist and the Bayesian. Frequentist approaches \cite{fox2011distinguishing} view the notion of probability by examining the statistical outcomes of repeated experiments in order to quantify the frequency of certain events, and perform maximum likelihood estimation to calibrate the parameters of a statistical model, while uncertainty is quantified by calibrated confidence intervals. On the other hand, Bayesian approaches \cite{lele2020should} are grounded on the laws of probability theory and provide a general formalism for posterior uncertainty quantification based on blending observational data and prior beliefs. Uncertainty in this case is obtained by performing posterior inference using Bayes' rule. By now, uncertainty quantification has grown into an active and mature field of research with diverse applications in computer vision \cite{kendall2017uncertainties}, language modeling \cite{xiao2019quantifying}, solutions of partial differential equations \cite{raissi2018hidden, yang2019adversarial,yang2018physics}, black-box function optimization \cite{blanchard2020bayesian}, sequential decision making \cite{yang2021output}, and engineering design \cite{sarkar2019multifidelity}. However, there are only few works \cite{psaros2022uncertainty,lin2021accelerated, moya2022deeponet} that have formulated uncertainty quantification methods in the context of operator learning.

In this work, we put forth a simple and effective framework for uncertainty quantification in operator learning, coined as {\em UQDeepONet}. Specifically, we propose a framework that combines an ensemble of DeepONet architectures together with randomized prior networks \cite{osband2018randomized, ciosek2019conservative} to control the model’s behavior in regions of the input space where training data does not exist. The main contributions of this work can be summarized as follows,
\begin{itemize}
    \item \textbf{Robust output function predictions:} By training an vectorized ensemble of randomized prior DeepONet, the proposed framework provides predictions for never-seen-before input functions with a small error spread compared to deterministic DeepONet.
    \item \textbf{Reliable uncertainty estimates:} Leveraging DeepONet ensembles in conjunction with randomized prior networks, the proposed model provides reliable uncertainty estimates for scarce data-sets that may include output functions with diverse magnitudes. 
    \item \textbf{Identification of out-of-distribution function samples:} Due to its ensemble structure, the proposed model can identify out-of-distribution data.
    \item \textbf{Handling noisy data:} The proposed model can be trained on a noisy data-set and provide meaningful predictions and uncertainty estimates for a never-seen before noisy data-set. 
\end{itemize}
Moreover, we provide a highly efficient software package for the vectorized training of UQDeepONets, enabling scalability to large data-sets, model architectures and ensemble sizes. The proposed implementation can be also trivially parallelized to leverage multiple hardware accelerators towards opening the path to performing reliable uncertainty quantification in realistic large-scale operator learning applications.

The paper is structured as follows. In section \ref{sec:Methods}, we review the key parts constructing this method. In particular, in section \ref{sec:DeepONet}, we present deep operator networks (DeepONet) through the lens of a simple example to point out some of its existing drawbacks. In section \ref{sec:UQDeepLearning}, we give a brief review of deep learning techniques for uncertainty quantification and introduce the randomized prior method. In section \ref{sec:UQDeepONet}, we present the details regarding the construction of the proposed UQDeepONet framework, and provide a pedagogical example to demonstrate the effectiveness of UQDeepONets. In section \ref{sec:Results}, three synthetic examples including anti-derivative of ordinary differential equations, reaction-diffusion equations and Burgers' equations and a realistic climate modeling example are presented to demonstrate and examine the capabilities of the proposed methodology. In section \ref{sec:discussion}, we present a discussion about our conclusions and some remarks.

\section{Methods}\label{sec:Methods}

\subsection{Deep operator networks}\label{sec:DeepONet}

\paragraph{Formulation:} Given $\curlyX \subset \R^{d_x}$ and $\curlyY \subset \R^{d_y}$ we refer to $x \in \curlyX$ as the input and $y \in \curlyY$ as the query locations, respectively. We use $C(\curlyX ; \R^{d_u})$ and $C(\curlyY ; \R^{d_s})$ to denote the spaces of continuous input and output functions, $u: \curlyX \to \R^{d_u}$ and $s: \curlyY \to \R^{d_s}$, respectively. Our goal is to learn an operator $\curlyF: C(\curlyX ; \R^{d_u}) \to C(\curlyY ; \R^{d_s})$ from pairs of input and output functions $\{ u^l, s^l \}$ generated from a ground truth operator $\curlyG: C(\curlyX ; \R^{d_u}) \to C(\curlyY ; \R^{d_s})$, where $u^l \in C(\curlyX ; \R^{d_u})$,  $s^l \in C(\curlyY ; \R^{d_s})$ and $l = 1, ..., N$, such that:
\begin{equation}\label{equ:NonlinearOperator}
    s^l = \curlyF(u^l).
\end{equation}
The theoretical basis for the construction of the DeepONet architecture is the universal approximation theorem of Chen \textit{et. al.} \cite{chen1995universal} for one hidden layer neural networks, which was later generalized to deep neural networks by Lu {\it et al.} \cite{lu2021learning}. The DeepONet architecture parametrizes the target operator $\curlyG(u)(y)$ using two neural networks, the branch network that receives an input function $u^l$, and the trunk network that receives a query location $y_j^l$. The $n$-dimensional features extracted by these two sub-networks are then merged by a dot-product operation to predict the value of the output function $s(y) = \curlyF_{\theta}(u)(y)$, given an input function $u$ and a query location $y$. In summary, the DeepONet representation of the latent operator can be expressed as
\begin{equation}\label{equ:DeepONet_forward_pass}
    \curlyF_{\theta} (u) (y) = \sum \limits_{i=1}^n b_{i}(u)  t_{i}(y),
\end{equation}
$b: \R^{m \times d_u} \to \R^{n \times d_s}$ denotes the branch network,  $t: \R^{1 \times d_y} \to \R^{n \times d_s}$ denotes the trunk network, and $\theta$ denotes all the trainable parameters of the branch and trunk networks, respectively. 

\paragraph{Training:}
Given a data-set consisting of input/output function pairs  $\{u^l, s^l \}_{l=1}^N$, a DeepONet model can be trained via empirical risk minimization using the following loss
\begin{equation}\label{equ:DeepONet_loss}
    \mathcal{L}(\theta) = \frac{1}{NP}\sum_{i=1}^{N}\sum_{j=1}^P|\curlyF_{\theta}(u^i)(y_j^i) - s^i(y_j^i)|^2.
\end{equation}
%
Recent works \cite{wang2021improved, di2021deeponet} have indicated that DeepONets trained using the loss of equation \eqref{equ:DeepONet_loss} can be biased towards approximating output functions with larger magnitudes. This is because output functions with larger magnitudes can dominate the magnitude of the back-propagated gradients during training, leading to a model that performs poorly when the magnitude of the queried output functions is small. To mitigate this shortcoming,  one can instead employ a normalized loss function, such as the one proposed in \cite{di2021deeponet, lu2021comprehensive}. Here we choose  
to normalize the loss by the infinity norm of the output function samples as 

\begin{equation}\label{equ:DeepONet_lossNorm}
    \mathcal{L}(\theta) = \frac{1}{NP}\sum_{i=1}^{N}\sum_{j=1}^P\frac{|\curlyF_{\theta}(u^i)(y_j^i) - s^i(y_j^i)|^2}{\|s^{i}(y^i_j) \|_{\infty}^2}.
\end{equation}
This normalized loss function weighs the different loss components by the magnitude of $s$, which has been shown \cite{wang2021improved, di2021deeponet} to safeguard the neural network from over-fitting low magnitude examples.

\paragraph{Inference:} Once a DeepONet model has been successfully trained, it can be queried to return predictions of a target output function at any continuous query location $y^\ast$, given an input function $u^\ast$. This can be translated to evaluating the model's forward pass as
\begin{equation}\label{equ:DeepONet_pred}
    \curlyF_{\theta}(u^\ast)(y^\ast) = \sum \limits_{i=1}^n b_i(u^\ast) t_i(y^\ast),
\end{equation}
where $u^*$ should be evaluated at the same locations $x$ as the input functions used during training, while $y^*$ are arbitrary query locations $y^* \in \mathcal{Y}$.

\paragraph{A need for uncertainty quantification:}\label{sec:AntiDerivative}
We consider the simple example of learning an anti-derivative operator to demonstrate some inherent shortcomings in the training of deterministic DeepONets that motivate the need for uncertainty quantification. To this end, let us consider a simple ordinary differential equation of the form
\begin{equation}\label{equ:Antiderivative_equ}
\begin{aligned}
    \frac{d s}{d x} &= u(x), \quad\quad\quad x \in  [0,1], \\
    s(0) &= 0.
\end{aligned}
\end{equation}
Our goal here is to learn the anti-derivative operator $\curlyG: u(x) \longrightarrow s(x)=s(0)+\int_{0}^{x} u(t) d t$ that maps a forcing term $u$ to the solution trajectory $s$. To do so, we create the training data-set by first sampling $u(x)$ from Gaussian process priors with different output scales that are chosen as $k = 10^{\alpha}$, where $\alpha$ is sampled from a uniform distribution $\mathcal{U}(-2, 2)$. For each output scale, we generate $100$ realizations of $u(x)$ and numerically integrate equation \eqref{equ:Antiderivative_equ} to create a total of $1,000$ training input/output function pairs. The test data-set is also created in an identical manner. We employ the Xavier normal initialization scheme \cite{glorot2010understanding} to initialize all neural networks weights, while all biases are initialized to zero. The resulting DeepONet model is trained using the Adam optimizer \cite{kingma2014adam} with default settings, using a learning rate of $10^{-3}$ with an exponential decay applied every $1,000$ iterations with rate of $0.9$. The predictive accuracy of the trained model is assessed using the relative $\mathcal{L}_2$ error defined as

\begin{equation}\label{equ:relative_2_error}
\begin{aligned}
    \mathcal{L}_2 &= \frac{\|\curlyF_{\theta}(u^\ast)(y^\ast) - s\|_2}{\|s\|_2}. \\
\end{aligned}
\end{equation}

\begin{figure}
\centering
\includegraphics[width=0.9\textwidth]{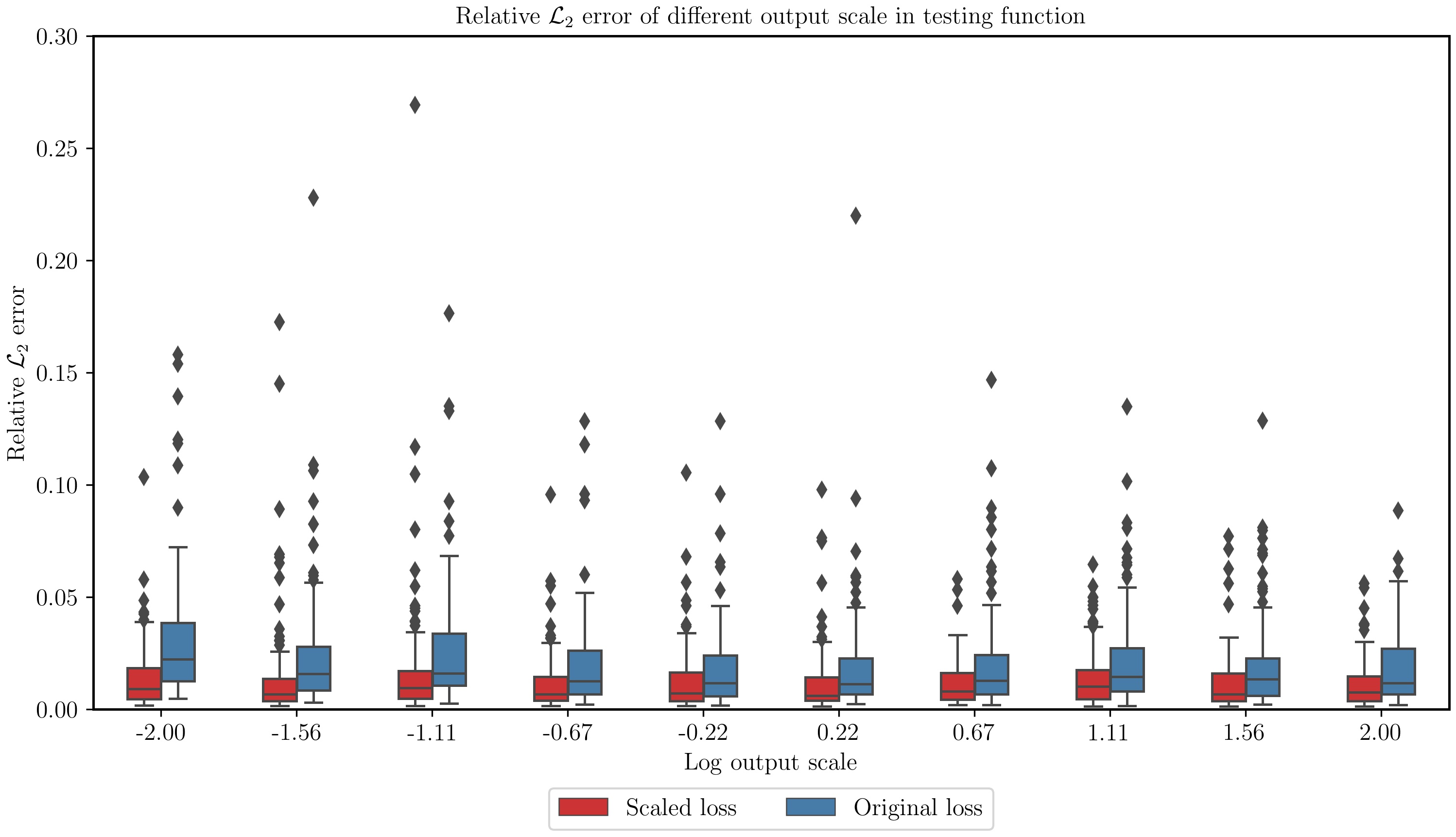}
\caption{{\em Learning the anti-derivative operator:} Comparison of relative $\mathcal{L}_2$ prediction errors of a DeepONet trained using the un-scaled loss of equation \eqref{equ:DeepONet_loss} (blue) and the scaled loss of equation \eqref{equ:DeepONet_lossNorm} (red). The black diamonds represent the error outliers among $100$ examples in the test data-set containing output functions with different output scales. x-axis is the $\alpha$ that denotes the log of the output scale. y-axis is the relative $\mathcal{L}_2$ prediction errors.}
\label{fig:Anti_Errors}
\end{figure}

In figure \ref{fig:Anti_Errors}, we report the comparison of the relative $\mathcal{L}_2$ prediction error averaged over all examples in the test data-set, for two different cases. In the first case, the model is trained using the loss in equation \eqref{equ:DeepONet_loss}, while in the second case we train the same model using the scaled loss in equation \eqref{equ:DeepONet_lossNorm}. Notice how the model trained using the loss of equation \eqref{equ:DeepONet_loss} yields larger errors compared to the model training using the scaled loss of equation \eqref{equ:DeepONet_lossNorm}. Second, the relative $\mathcal{L}_2$ errors of the model trained with the un-scaled loss increase as the output scale of the target functions decreases. Third, the maximum error across all testing data is around $25\%$, which indicates that the worst-case-scenario prediction of DeepONets could be unreliable. As summarized in table \ref{tab:error_magnitude}, there is a clear trend of increasing errors as the output scale of the target output functions function is decreased, suggesting that DeepONets can be biased towards approximating function with large magnitudes -- an observation that is in agreement with the findings reported in \cite{di2021deeponet,wang2021improved} . These findings demonstrate some drawbacks that are inherent to the deterministic training of DeepONets, and motivate the need for an uncertainty quantification method. 

\begin{table*}[!ht]
    \centering
    \caption{{\em Learning the anti-derivative operator:} Relative $\mathcal{L}_2$ error averaged over groups of examples in the test data-set with the same output scale, for a DeepONet trained using the un-scaled loss of equation \eqref{equ:DeepONet_loss}.}
    \label{tab:error_magnitude}
\resizebox{\textwidth}{!}{%
    \begin{tabular}{l|cccccccccc}
      \toprule 
        Output magnitude in $\log$-scale & \bfseries -2.00 & \bfseries -1.56 & \bfseries -1.11 & \bfseries -0.67 & \bfseries -0.22 & \bfseries 0.22 & \bfseries 0.67 & \bfseries 1.11 & \bfseries 1.56 & \bfseries 2.00\\ 
      \midrule 
      Average relative $\mathcal{L}_2$ error (\%) & 3.07 & 2.59 & 2.85 & 2.09 & 1.86 & 1.83 & 2.08  & 2.35 & 1.94 & 1.81 \\
      \bottomrule 
    \end{tabular}
}
\end{table*}

\subsection{Uncertainty quantification in deep learning}\label{sec:UQDeepLearning}

Uncertainty quantification in deep learning has recently received a lot of attention and has been successfully employed for diverse applications, including computer vision \cite{kendall2017uncertainties}, language modeling \cite{xiao2019quantifying}, solutions of partial differential equations \cite{raissi2018hidden, yang2019adversarial}, black-box function optimization \cite{snoek2015scalable} and sequential decision making \cite{osband2018randomized, riquelme2018deep}. In this section, we provide a brief review of the most widely used tools for performing uncertainty quantification in deep neural networks. A more detailed overview can be found in \cite{psaros2022uncertainty}, while extensive comparisons between different methods have been recently performed in \cite{osband2021evaluating, ciosek2019conservative}. 

\paragraph{Hamiltonian Monte Carlo:} Hamiltonian Monte Carlo (HMC) methods \cite{neal2012bayesian} are currently considered the gold-standard tool for approximate Bayesian inference with asymptotic convergence guarantees. They are formulated on the premise of exploiting gradient information and trajectory dynamics in the parameter space for accelerating the convergence of a Markov Chain Monte Carlo sampling process. Despite its success across many applications in machine learning, the use of HMC in deep learning is typically limited by its high computational cost and its requirement for using the entire training data set for evaluating the likelihood function and its gradients during the Metropolis accept-reject step \cite{yang2021b}. These factors cast a great difficulty in scaling HMC algorithms to large neural network architectures and large data-sets.

\paragraph{Variational inference:} Variational inference \cite{hinton1993keeping} utilizes tractable distributions (e.g. the mean field family) to approximate intractable posterior distributions by enabling the derivation of a tractable optimization objective, such as the KL-divergence, for updating a model's parameters via stochastic gradient descent methods. Variational inference can be extended for estimating the posterior distribution of large neural networks \cite{blundell2015weight}, but mean-field approximations are known to underestimate the posterior variance, while they also provide only a crude approximation of the degenerate and multi-modal posterior distributions often encountered in Bayesian deep learning. As such, the resulting uncertainty estimates are often poor and lead to models that under-perform in settings where well-calibrated uncertainty estimation plays a prominent role (e.g. sequential decision making \cite{riquelme2018deep, osband2021evaluating}).

\paragraph{Deep Ensembles:} Deep ensembles \cite{lakshminarayanan2016simple} provide a straightforward frequentist approach, in which independent neural networks with different initialization are trained in parallel. Such approach aims to calibrate the epistemic uncertainty of neural networks whose training objective is non-convex and might involve many local optima leading to multi-modal posterior distributions that are hard to explore via sampling \cite{fort2019deep}. The deep ensemble method is shown to produce well-calibrated uncertainty estimates \cite{lakshminarayanan2016simple}, while it can readily facilitate vectorization and parallelization in modern hardware accelerators, enabling uncertainty quantification for large deep learning models and large data-sets within a reasonable computational cost. This approach was successfully adopted in \cite{psaros2022uncertainty} for performing uncertainty quantification in DeepONets, although it was limited to using small ensembles with 5-10 replicas, likely due to a high computational cost.

\paragraph{Dropout:} Dropout \cite{srivastava2014dropout} is a simple trick that randomly zeros out the output of neurons during training and inference based on a fixed a-priori chosen probability. Dropout does not only work as a simple regularization mechanism, but also enables the quantification of uncertainty in neural networks. It was also shown in \cite{gal2016dropout} that in deep Gaussian processes \cite{rasmussen2004gaussian}, where the width of the neural networks tends to infinity, dropout works as an approximation of Bayesian inference. Dropout has a very low computational overhead and has been widely used across different deep learning architectures and applications, however it is known to provide crude uncertainty estimates that often lead to models that under-perform in settings where well-calibrated posterior uncertainty plays a prominent role (e.g. sequential decision making \cite{riquelme2018deep, osband2021evaluating}).

\paragraph{Stochastic Weight Averaging:}
Stochastic Weight Averaging (SWA) \cite{izmailov2018averaging} is an ensembling procedure that averages multiple model parameters along the training trajectory of a deep learning model. Using a cyclical learning rate, stochastic gradient descent has the ability to better explore the underlying non-convex loss landscape, leading to stochastic weights that correspond to different local basins of attraction. This framework has been demonstrated to yield robust predictions and achieve generalizable solutions with low loss. Based on SWA, SWA-Gaussian (SWAG) \cite{maddox2019simple} was proposed for conducting uncertainty representation and Bayesian inference in deep networks. SWAG models the model parameters as samples of multivariate Gaussian distributions. Then the SWA solution as considered as the posterior mean, while a low-rank approximation is proposed to efficiently estimate the posterior covariance. This framework has shown benefits in performing out-of-distribution detection and provides robust prediction. However, although the low-rank approximation is used for the covariance matrix, it is still computationally expensive to draw posterior samples for large neural network architectures.

\paragraph{Randomized prior networks:} Randomized prior networks were proposed by Osband {\it et al.} \cite{osband2018randomized} in the context of leveraging posterior uncertainty estimates to effectively balance the exploration versus exploitation trade-off in reinforcement learning. The main idea consists of using randomly initialized networks to control the model’s behavior in regions of the space where limited training data is available. Compared to deep ensembles \cite{lakshminarayanan2016simple}, this method introduces randomly initialized prior functions whose parameters remain fixed during model training, and sums the output of those fixed prior functions with the output of a trainable neural network. To this end, a prediction can be obtained by evaluating
\begin{equation}\label{equ:randomized_prior}
    \hat{f}_{\theta} = f_{\theta} + \beta f_{\Theta},
\end{equation}
where $f_{\theta}$ is a neural network with trainable parameters $\theta$, while $f_{\Theta}$ denotes the same exact network that is randomly initialized with another set of parameters $\Theta$ that remained fixed during training. A frequentist ensemble for uncertainty quantification can be constructed by considering multiple replicas of randomly initialized prior networks. During training, one keeps the parameters of each prior network $f_{\Theta}$ in the ensemble fixed, and only updates the trainable parameters $\theta$.
The fixed neural networks $f_{\Theta}$, represent our prior belief and act as the source of uncertainty. Moreover, $\beta$ is a user-defined hyper-parameter that controls the variance of the uncertainty originating from the prior networks. This approach is supported by theoretical guarantees of consistency and asymptotic convergence to the true posterior distribution  in the context of Bayesian inference for Gaussian linear models \cite{osband2018randomized}, and has been recently demonstrated to be very effective in complex sequential decision making tasks \cite{osband2021evaluating, ciosek2019conservative}. Finally, similar to deep ensembles, randomized prior networks can be trivially parallelized allowing one to accommodate large network architectures and training data-sets.

\paragraph{Does Bayesian deep learning work?}
High quality uncertainty estimation plays critical roles in many applications of Bayesian deep learning. Recent work \cite{osband2021evaluating} argues that despite the success of Bayesian deep learning frameworks for approximating marginal distributions, some established methods yield poor performance when evaluating joint predictive distributions. For instance, the authors argue that variational inference, dropout and deep ensembles fail to provide reliable uncertainty estimates and show poor performance in areas where uncertainty quantification plays essential roles, such as sequential decision making tasks. HMC provides state-of-the-art performance in estimating uncertainty, but typically suffers from a high computational cost that limits its applicability to large-scale problems. Randomized priors, on the other hand, have been shown to achieve similar performance to HMC, which is especially beneficial in the small data regime, while being computationally efficient and trivially parallelizable on hardware accelerators. Moreover, a recent study \cite{ciosek2019conservative} provides theoretical results showing that the uncertainty obtained from randomized prior networks is conservative and capable of reliably performing out-of-distribution sample detection. The same study also shows that this framework generalizes to any neural network architecture trained by standard optimization pipelines. Such properties make the randomized prior approach a potentially valuable tool for applying Bayesian deep learning methods in challenging engineering problems.


\paragraph{Other related works:} 
In \cite{lin2021accelerated, moya2022deeponet}, the authors proposed a Markov Chain Monte Carlo (MCMC) approach for posterior inference in DeepONet architectures. Although MCMC techniques are considered to be the gold standard for posterior uncertainty quantification of complex models, they typically suffer from high computational cost and cannot be easily scaled to efficiently support large model architectures and large data-sets. In \cite{moya2022deeponet}, the authors also provided a maximum likelihood framework to train DeepONet. They conducted the uncertainty quantification with a Gaussian estimate of the predictive distribution. In \cite{psaros2022uncertainty}, the authors provided a comprehensive review of uncertainty quantification methods, including some early results on uncertainty quantification in DeepONet. On one hand, they focus on facilitating aleatoric uncertainty estimation on noisy data. To this end, the authors use pre-trained DeepONet on noiseless data as a surrogate, and quantify the uncertainty from random input functions through posterior inference methods such as Bayesian neural networks (BNN) \cite{neal2012bayesian} or Generative adversarial nets (GAN) \cite{goodfellow2014generative}. However, these approaches require the pre-training data to be noiseless, while the computational cost for training BNNs and GANs can be high. On the other hand, similar to deep ensemble \cite{lakshminarayanan2016simple}, \cite{psaros2022uncertainty} also considers a frequentist approach by training a small number of neural networks in order to capture epistemic uncertainty in the DeepONet model predictions. Although this is a computationally efficient approach, the findings discussed in \cite{ciosek2019conservative, osband2021evaluating} indicate that deep ensembles can yield sub-optimal performance by providing overconfident uncertainty estimation in practice. In particular, \cite{ciosek2019conservative} points out an edge case, in which if each of the networks in the ensemble is convex in the weights, all networks would converge to the same weights and therefore result in zero uncertainty. 
To overcome the aforementioned difficulties, in the next section we put forth a scalable and computationally efficient framework for conservative posterior uncertainty quantification in DeepONet, by formulating a novel DeepONet variant that leverages the use of randomized prior networks \cite{osband2016deep}.

\subsection{Uncertainty quantification in DeepONet using randomized priors}\label{sec:UQDeepONet}

\paragraph{Formulation:} Leveraging the DeepONet architecture for operator learning and the flexibility of randomized prior networks for uncertainty quantification, we propose a novel model class called {\em UQDeepONet} that facilitates scalable uncertainty estimation in operator learning. Using the conventional DeepONet parametrization of equation \eqref{equ:DeepONet_forward_pass} as a starting point, we will parameterize the branch and the trunk networks, $b(u)$ and $t(y)$, respectively, using fully connected architectures, and then augment the model with randomized priors
\begin{align}
    \hat{b}_{\phi}(u) &= b_{\phi}(u) + \beta b_{\Phi}(u), \label{equ:UQDeepONet_branch_trunk_1}\\
    \hat{t}_{\psi}(y) &= t_{\psi}(y) + \beta t_{\Psi}(y), \label{equ:UQDeepONet_branch_trunk_2}
\end{align}
where $b_{\Phi}(u)$ and $t_{\Psi}(y)$ denote randomly initialized branch and trunk networks with the exactly the same architecture as the backbone trainable networks $\hat{b}_{\phi}(u)$ and $\hat{t}_{\psi}(y)$, albeit with a different set of parameters $\Phi$ and $\Psi$ that are randomly initialized and remain fixed during model training. Under this setup, the prediction of a UQDeepONet model is obtained as
\begin{equation}\label{equ:UQDeepONet_pred}
\begin{aligned}
    \curlyF_{\theta}(u^*)(y^*) = \sum \limits_{i=1}^n \hat{b}_{\phi}(u^*)_i \hat{t}_{\psi}(y^*)_i,  \\
\end{aligned}
\end{equation}
where $\theta = [\phi, \psi]$ denotes the total number of trainable parameters of the UQDeepONet, which does not include the prior network parameters $\Phi$ and $\Psi$.
Similar to \cite{osband2018randomized}, ensemble predictions can be obtained by considering multiple replicas of randomly initialized prior networks. During training, one keeps the $\Phi$ and $\Psi$ parameters of each prior network in the ensemble fixed, and only updates the trainable parameters $\theta$.
Notice that the forward evaluation of the model is decoupled across ensembles; a fact that readily facilitates an implementation that is trivially parallelizable. As we will later demonstrate, this enables the concurrent training of hundreds of DeepONet on modern GPU accelerators with minimal computational overhead.  Once a UQDeepONet is trained, we can obtain statistical estimates for the model's predictions and variance via Monte Carlo ensemble averaging
\begin{align}
    \mu (s|u^*,y^*) & = \mathbb{E}_{p_s}[s|u^*,y^*] \approx \frac{1}{N_s}\sum_{i=1}^{N_s} \Big ( \curlyF_{\theta}(u^*)(y^*) \Big )_i, \label{equ:predictive_mean} \\
    \sigma^{2}(s|u^*,y^*) & = \mathbb{V}\text{ar}_{p_{s}}[s|u^*,y^*] \approx \frac{1}{N_s}\sum_{i=1}^{N_s} \Big [ \Big( \curlyF_{\theta}(u^*)(y^*) \Big)_i - \mu(s|u^*,y^*)  \Big ]^2, \label{equ:predictive_variance}
\end{align}
where, $N_s$ is the number of DeepONet in the UQDeepONet ensemble. The prediction and uncertainty estimation can be obtained by evaluating the forward pass of each network in the ensemble, which is also trivially parallelized and computationally efficient.

\begin{figure}
\centering
\includegraphics[width=0.9\textwidth]{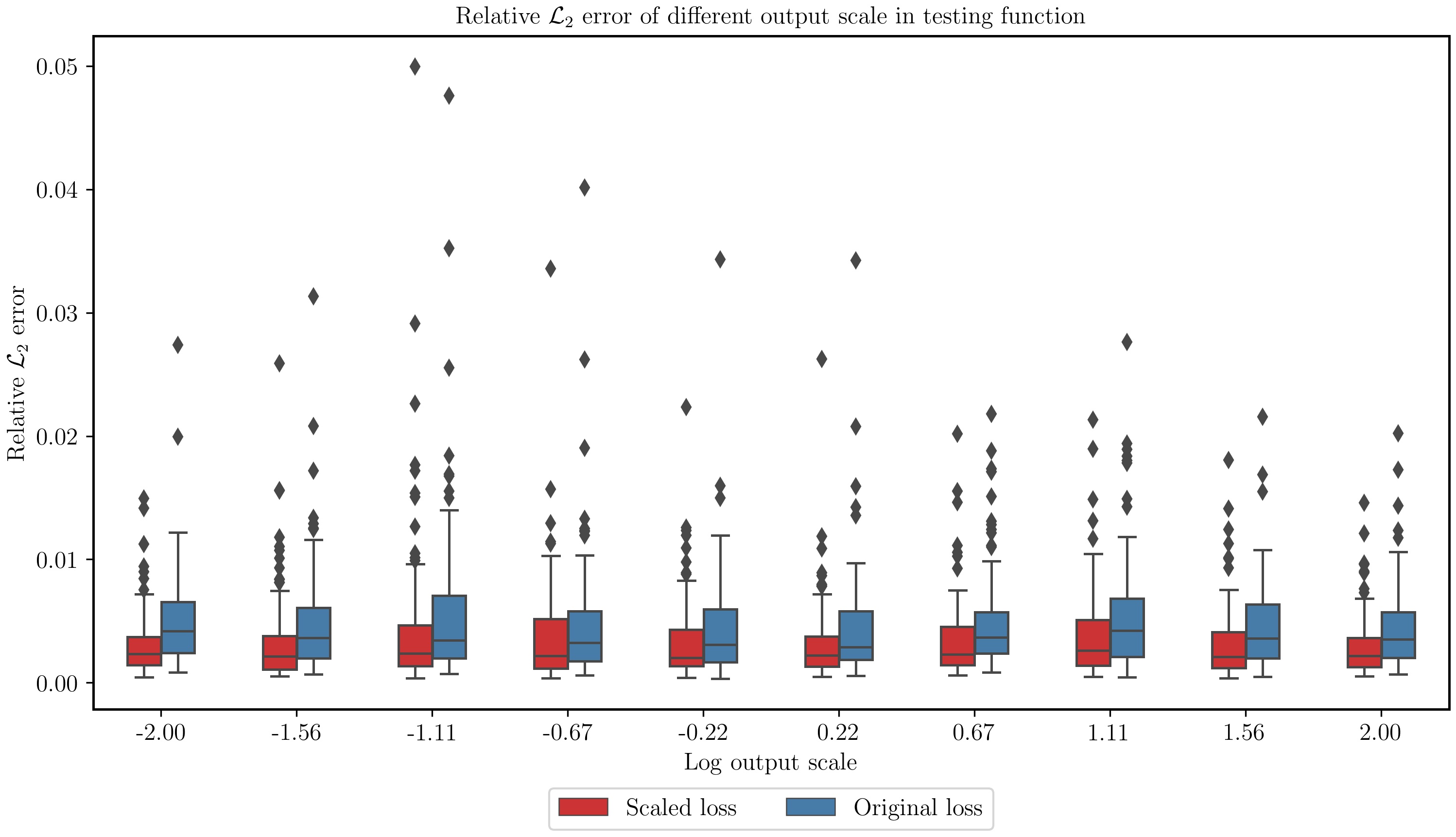}
\caption{{\em Learning the anti-derivative operator:} Comparison between the relative $\mathcal{L}_2$ error of the predictions of a UQDeepONet trained using the un-scaled loss  of equation \eqref{equ:DeepONet_loss} (blue) and the scaled loss of equation \eqref{equ:DeepONet_lossNorm} (red). The black diamonds represent the error outliers among $100$ examples in the test data-set containing output functions with different output scales.  x-axis is the $\alpha$ that denotes the log of the output scale. y-axis is the relative $\mathcal{L}_2$ prediction errors.}
\label{fig:Anti_Errors_UQDeepONet}
\end{figure}

\begin{figure}
\centering
\includegraphics[width=\textwidth]{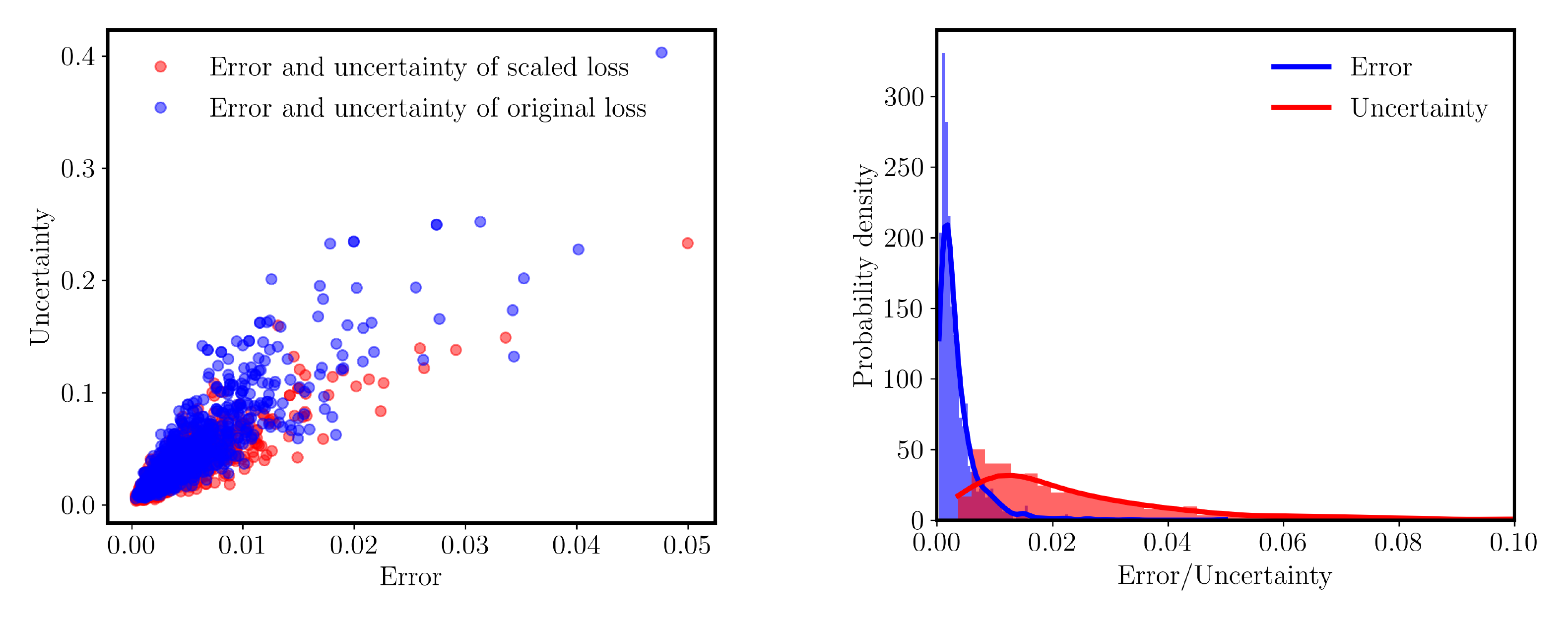}
\caption{{\em Uncertainty quantification versus prediction errors:} {\it Left:} the $\curlyL_2$ errors, scaled $\curlyL_2$ errors and scaled uncertainty of the UQDeepONet predictions for different output scales are presented. The blue dots represent the $\mathcal{L}_2$ error versus the quantified uncertainty for the original loss in equation \eqref{equ:DeepONet_loss}. The red dots represent the $\mathcal{L}_2$ error versus the quantified uncertainty for the scaled loss in equation \eqref{equ:DeepONet_lossNorm}. {\it Right:} Marginal density of prediction error (red) and predictive uncertainty (blue).}
\label{fig:Anti_Error_vs_Uncertainty}
\end{figure}

\paragraph{Learning the anti-derivative operator with quantified uncertainty:}
To provide a simple use case of the proposed UQDeepONet framework, let us revisit the anti-derivative example discussed in section \ref{sec:AntiDerivative}. To this end, we will approximate the anti-derivative operator with a UQDeepONet model and report two metrics for assessing the model performance. Similar to the definition of relative $\mathcal{L}_2$ loss in equation \eqref{equ:relative_2_error}, the relative uncertainty defined as 

$$\textrm{uncertainty}_2 = \frac{\|\sigma (s|u^*, y^*)\|_2}{\|s\|_2},$$

where $\sigma (s| u^*, y^*)$ is defined in equation \eqref{equ:predictive_variance}. The proposed scaling of the uncertainty metric is tailored to report higher uncertainty in areas where the prediction  error is higher. The error plot is presented in figure \ref{fig:Anti_Errors_UQDeepONet}. The UQDeepONet model is trained using Adam optimizer \cite{kingma2014adam} with learning rate $l_r = 10^{-3}$, which we decrease every $1,000$ iterations by $0.9$. The total number of iterations is $5\times 10^{5}$. Similar to figure \ref{fig:Anti_Errors}, we observe that the errors resulting from using the scaled $\mathcal{L}_2$ loss for training are smaller than the errors from $\mathcal{L}_2$ loss. When comparing outlier cases, the predictions of the UQDeepONet show a significantly smaller error spread than a conventional DeepONet providing a maximum error around $5\%$ in most cases, while the error spread is also relatively contained across different output scale magnitudes. Moreover, in figure \ref{fig:Anti_Error_vs_Uncertainty}, we demonstrate the uncertainty estimation obtained from the UQDeepONet, which further reflects this point by providing higher uncertainty for test examples with higher prediction error. We also report the marginal density of prediction error versus uncertainty estimates generated by the UQDeepONet. As one would hope for, the predictive uncertainty attains small values for the cases where the model predictions are accurate, while it also consistently identifies cases for which the prediction error is large. This is a good indication that UQDeepONet is able to provide calibrated and conservative uncertainty estimates.

\paragraph{Parallel implementation, computational cost and sensitivity on the $\beta$ parameter:} 
A discussion on parallel implementation and its computational cost is provided in the supplementary material (see section \ref{sec:parallel_cost_supp}). Specifically, we report wall-clock time needed for the training of a UQDeepONet with different ensemble sizes, and observe that the proposed parallel implementation is computationally efficient as the cost exhibits a near-linear scaling with the number networks in the ensemble. 

We also conduct a study on how the different values of the $\beta$ hyper-parameter in the randomized prior networks affect the model output (see equations \eqref{equ:UQDeepONet_branch_trunk_1} and \eqref{equ:UQDeepONet_branch_trunk_2}). The discussion is provided in supplementary material section \ref{sec:anti_studybeta_supp}, indicating that the default choice of $\beta=1$ adopted in the randomized prior literature \cite{osband2018randomized, osband2021evaluating, ciosek2019conservative} yields robust performance.

\section{Results}
\label{sec:Results}

To demonstrate the performance of the proposed UQDeepONet framework we have conducted a series of numerical studies across four representative benchmarks. First, we revisit the anti-derivative example discussed in the previous sections to study the robustness of UQDeepONets. The next two benchmarks consider canonical cases of approximating the solution operator of partial differential equations, while the last benchmark is focused on learning a black-box operator for a climate modeling prediction task from real sensor measurements. In each case, we aim to demonstrate the ability of UQDeepONets to provide robust output function predictions, return reliable uncertainty estimates, as well as their ability to identify model bias, out-of-distribution function samples, and handle noisy data. Our objectives for each experiment can be summarized as follows.

\paragraph{Robust output function prediction:} We consider the anti-derivative example described in section \ref{sec:AntiDerivative} where our goal is to learn an operator that maps random forcing terms to the solution trajectory of an ordinary differential equation. We use this example to quantitatively demonstrate how UQDeepONets can provide robust output function predictions across different examples in the testing data-set.

\paragraph{Identification of out-of-distribution function samples:} We consider a reaction-diffusion system as an example of learning an operator that maps random source terms to the full spatio-temporal solution of a time-dependent partial differential equation. We use this example to empirically illustrate how UQDeepONets can provide reliable uncertainty estimates, allowing us to identify out-of-distribution examples in the testing data-set. 

\paragraph{Reliable uncertainty estimates:} We consider a time-dependent nonlinear transport problem with the goal of learning an operator that maps random initial conditions to the full spatio-temporal solution of the viscous Burgers equation. We choose this benchmark to show that UQDeepONets can provide both robust predictions and conservative uncertainty estimates, even for the worst-case-scenario in the testing data-set.

\paragraph{Handling noisy data:} We consider a large-scale climate modeling task where our goal is to learn an operator that maps the surface air temperature field over the Earth to the surface air pressure field, given real weather station data. Here we demonstrate the ability of UQDeepONet to handle noisy data, as well as to provide robust predictions and reliable uncertainty estimates, even when the model is called to extrapolate on future data. 

\paragraph{Hyper-parameter settings:} Across all benchmarks, we parametrize the branch and trunk networks using a multi-layer perceptron with 3 hidden layers and 128 neurons per layer. The randomized priors ensemble size for all examples is fixed to $N_s=512$. The details of neural networks architecture settings are summarized in table \ref{tab:NNarchitecture}. All models are trained on $N$ input/output function pairs $ \{ u^i, s^i \}$, and for each pair we assume that we have access to $m$ discrete measurements of the input function $\{ u^i_1, ..., u^i_m \}$, and $M$ discrete measurements of the output function $\{ s^i_1, ..., s^i_M \}$ at query locations $\{ y^i_1, ..., y^i_M \}$. A different subset of $P$ points is chosen for each DeepONet in the ensemble. We employ Glorot normal initialization \cite{glorot2010understanding} for all examples. The Adam optimizer is used to optimize all models with a base learning rate $l_r = 10^{-3}$, which we decrease every $1,000$ iterations by a factor of $0.9$. The total number of iterations is $5\times 10^{5}$. All hyper-parameter settings are summarized in table \ref{tab:TrainingSetting}. For all examples, we set the inputs and outputs dimensions as discussed in \ref{sec:UQDeepONet} and normalize the data using the infinity norm, as discussed in section \ref{sec:DeepONet}.

\paragraph{Training:} For training the UQDeepONets,  we propose a novel hierarchical algorithm for performing unbiased mini-batch sampling in a very efficient manner. At each training iteration, we randomly choose $N_s$ out of the $M$ total available labels which are shared across all DeepONets in the randomized priors ensemble. Then, we fetch different random batches (one for each DeepONet in the ensemble) of size $N_b$ out of the $N$ available input/output function pairs to create a data-set of size $N_b \times N_s \times d_s$. This approach significantly reduces the computational cost and memory footprint of the data-loader during training.  

\paragraph{Error and performance metrics:} For all examples, we report the relative $\curlyL_2$ error between the ground truth and the mean prediction of a UQDeepONet, as defined in equation \eqref{equ:relative_2_error}. All computations are vectorized on a single NVIDIA RTX A6000 GPU, leading to a training time of 6 hours for all benchmarks. Our implementation can also be readily extended to accommodate data-parallelism on multiple GPUs, which we anticipate that can considerably reduce the total training time in the future.

\begin{table*}[!ht]
    \centering
    \caption{{\em Hyper-parameter settings:} UQDeepONet architecture used across all benchmarks.}\label{tab:NNarchitecture}
    \begin{tabular}{l|cccccccccc}
      \toprule 
     Architecture & \bfseries  Trunk depth & \bfseries Trunk width & \bfseries Branch depth & \bfseries Branch width & \bfseries Ensemble size \\
      \midrule 
     Setting & \bfseries $3$ & \bfseries  $128$ & \bfseries $3$ & \bfseries  $128$ & \bfseries $512$ \\
      \bottomrule 
    \end{tabular}
\end{table*}

\begin{table*}[!ht]
    \centering
    \caption{{\em Hyper-parameter settings:} Training  settings used across all benchmarks.}
    \label{tab:TrainingSetting}
    \begin{tabular}{l|cccccccccc}
      \toprule 
     Hyper-parameter & \bfseries Learning rate & \bfseries Optimizer & \bfseries  Iterations & \bfseries Learning rate decay  \\ 
     \midrule 
      Setting  &  $10^{-3}$ & Adam \cite{kingma2014adam} &  $5\times 10^{5}$ & $0.9$ every 1,000 iterations \\
      \bottomrule 
    \end{tabular}
\end{table*}

\subsection{Robust output function prediction}\label{sec:Robustness_anti}
We revisit the anti-derivative example in section \ref{sec:AntiDerivative} to study the robustness of the UQDeepONet predictions by reporting the maximum test relative $\mathcal{L}_2$ error in table \ref{tab:robustness}, which is an indicator of the worst-case-scenario prediction. We find that the maximum error reduces significantly as the ensemble size increases. When the ensemble size is greater than $32$, the maximum testing error saturates to a limiting value. This observation demonstrates the ability of the proposed framework to reduce the inaccuracies for corner cases in the testing data-set, and yield robust worst-case-scenario predictions. Specifically, an ensemble with 32 networks yields a 6x reduction in relative error for the worst-case-scenario prediction compared to a conventional DeepONet model.

\begin{table*}[!ht]
    \centering
    \caption{{\em Robust output function prediction:} Maximum testing relative $\mathcal{L}_2$ error of a UQDeepONet for different ensemble sizes.}\label{tab:robustness}
\resizebox{\textwidth}{!}{%
    \begin{tabular}{l|cccccccccc}
      \toprule 
        Ensemble size & \bfseries 1 & \bfseries 2 & \bfseries 4 & \bfseries 8 & \bfseries 16 & \bfseries 32 & \bfseries 64 & \bfseries 128 & \bfseries 256 & \bfseries 512\\ 
      \midrule 
      Max relative $\mathcal{L}_2$ error (\%)  & 26.91 & 20.12 & 8.31 & 9.89 & 6.26 & 4.37 & 4.36  & 4.43 & 4.51 & 4.99 \\
      \bottomrule 
    \end{tabular}
}
\end{table*}

\subsection{Identification of out-of-distribution function samples}\label{sec:ReactionExamples}
\begin{figure}
\centering
\includegraphics[width=0.9\textwidth]{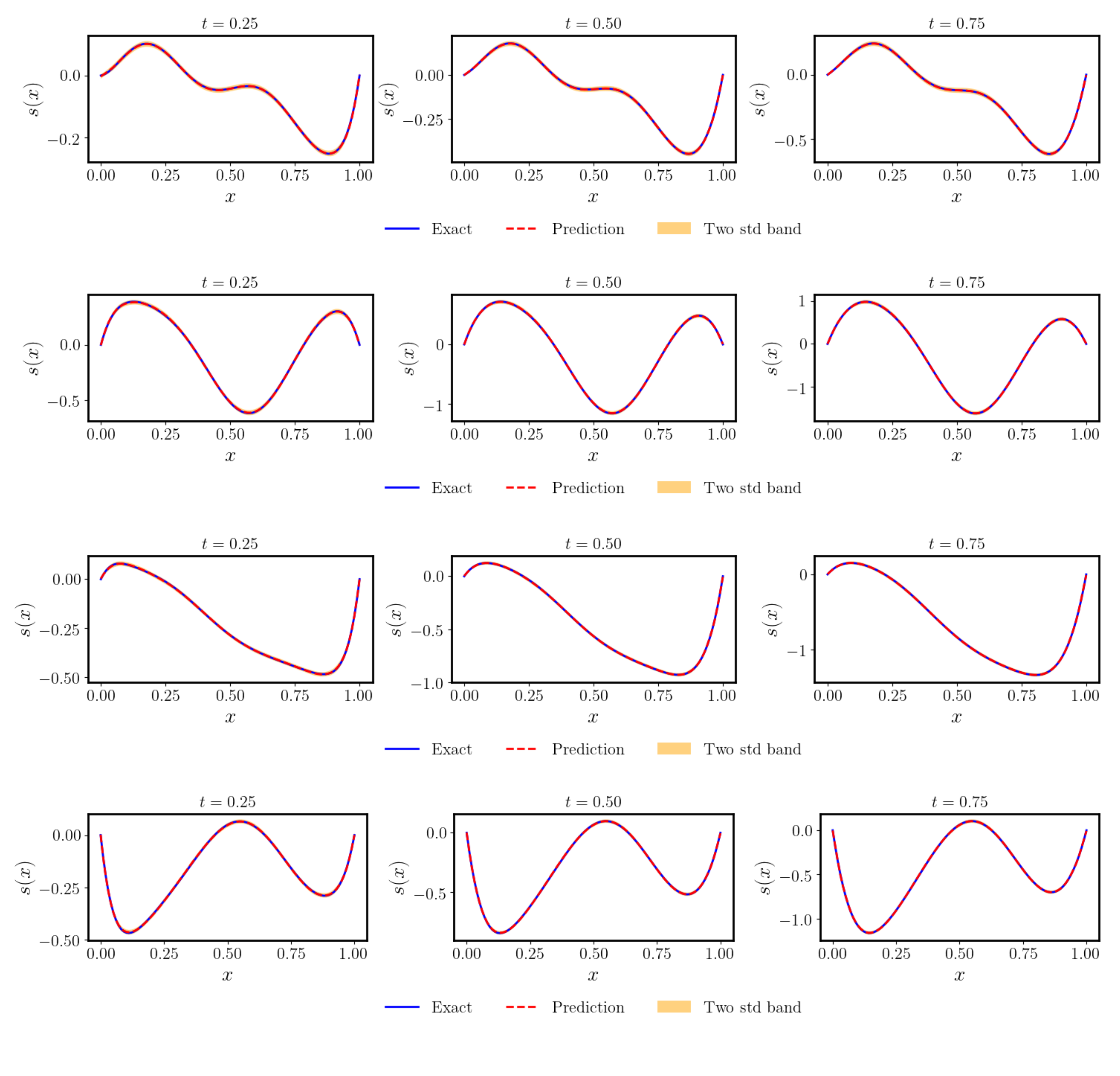}
\caption{{\em Reaction-diffusion dynamics:} UQDeepONet predictions for representative examples in the test data-set (top to bottom rows).  For each example, we present a comparison between the ground truth solution (blue), the predictive mean of the UQDeepONet (red dashed), and the predictive UQDeepONet uncertainty (yellow shade), for $t = 0.25$, $t = 0.50$ and $t = 0.75$.}
\label{fig:Reaction_realizations}
\end{figure}

Reaction-diffusion equations \cite{smoller2012shock} are widely used to describe the behavior of chemical reactions \cite{grzybowski2009chemistry}, pattern formation \cite{kondo2010reaction}, ecology \cite{cantrell2004spatial} and physics  \cite{smoller2012shock}. In one spatial dimension, they take the following form
\begin{equation}\label{equ:Reaction_equ}
\begin{aligned}
    \frac{\partial s}{\partial t} &= \nu\frac{\partial^2 s}{\partial x^2} + k s^2 + u(x), \quad\quad\quad (x, t) \in [0,1]\times [0,1], \\
    s(x, 0) &= 0,\quad\quad  x\in [0,1], \\
    s(0, t) &= 0,\quad\quad  t\in [0,1], \\
    s(1, t) &= 0,\quad\quad  t\in [0,1], \\
\end{aligned}
\end{equation}
where $\nu$ denotes the diffusion coefficient and $k$ the reaction rate, both of which are here set to $0.01$. The input and the output domain in this problem are $\curlyX$ and $\curlyY$, respectively, with $d_x =1$ and $d_y = 2$. Moreover, because we map scalar functions (i.e. $d_u = d_s = 1$) the operator we are approximating is $\curlyG: C(\curlyX; \R) \to C(\curlyX; \R)$. The training and testing data generation process for this example is summarized in the supplementary material (see section \ref{sec:data_creation_supp}). Here we present the UQDeepONet predictions for randomly chosen examples in the test data-set in figure \ref{fig:Reaction_realizations}. For most random sampled examples, we observe very accurate predictions with small prediction variance, indicating that the UQDeepONet model is confident in what it predicts.

\begin{figure}
\centering
\includegraphics[width=\textwidth]{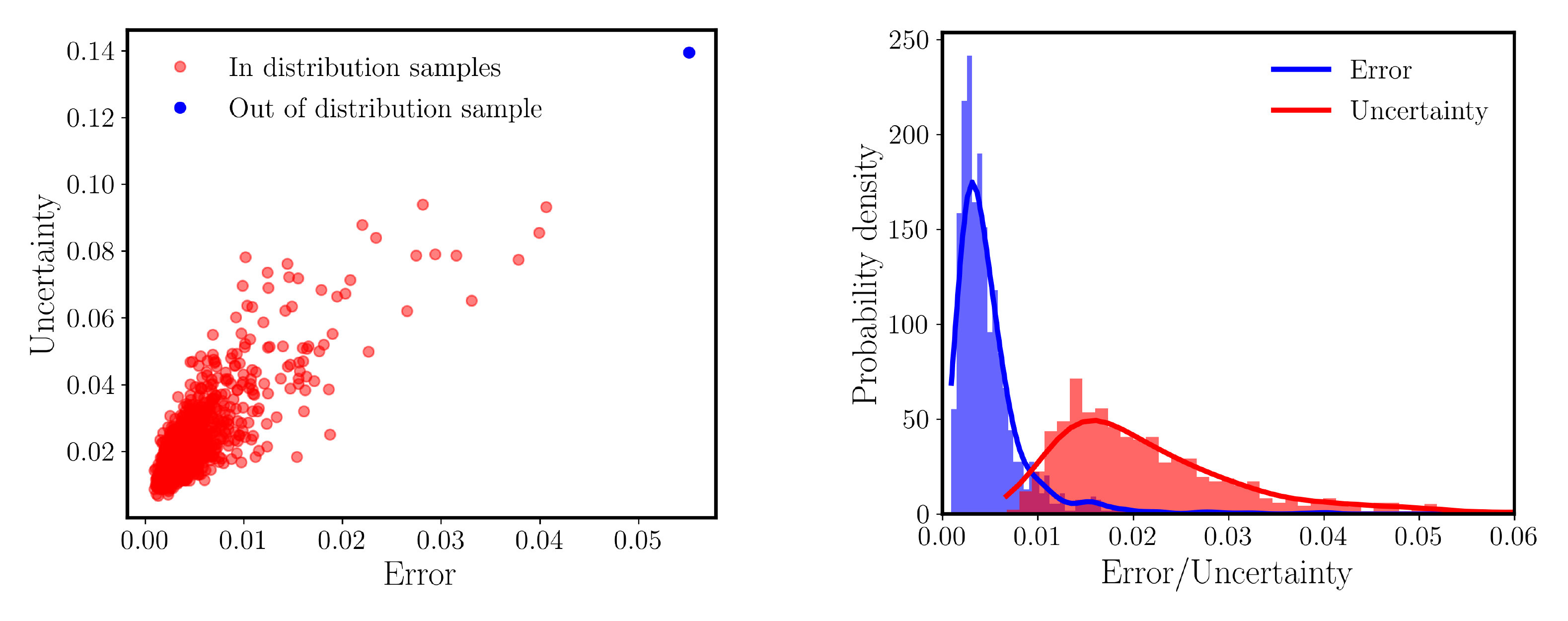}
\caption{{\em Reaction-diffusion dynamics:} {\it Left:} uncertainty quantification versus prediction errors. The scaled $\curlyL_2$ errors and scaled uncertainty of the UQDeepONet predictions are presented. The red dots represent the $\mathcal{L}_2$ error versus the quantified uncertainty for the scaled loss in equation \eqref{equ:DeepONet_lossNorm}. The blue dot represents an outlier corresponding to an out-of-distribution function sample in the testing data set. \ {\it Right:} Marginal density of prediction error (red) and predictive uncertainty (blue).}
\label{fig:Reaction_Error_vs_Uncertainty}
\end{figure}

\begin{figure}
\centering
\includegraphics[width=0.9\textwidth]{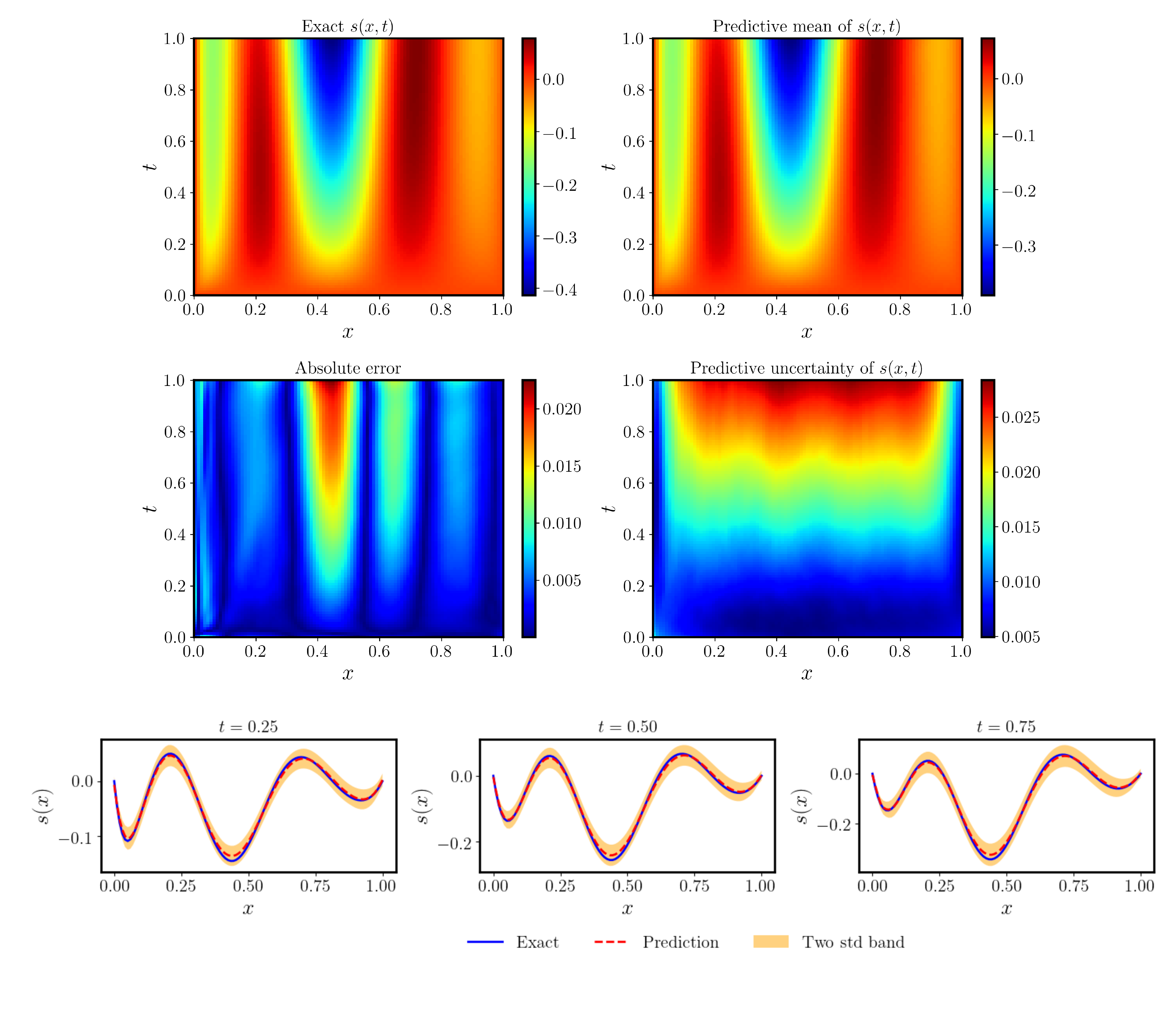}
\caption{{\em Reaction-diffusion dynamics:} Worst case scenario prediction. {\it Top panel:} Ground truth solution and UQDeepONet predictive mean. {\it Middle panel:} Absolute error between the ground truth solution and the UQDeepONet predictive mean, as well as the UQDeepONet predictive uncertainty. {\it Bottom panel:} Ground truth solution (blue), UQDeepONet predictive mean (red dashed) and uncertainty (yellow shade) at $t = 0.25$, $t = 0.50$ and $t = 0.75$.}
\label{fig:Reaction_max}
\end{figure}

We report the prediction error versus the quantified uncertainty of UQDeepONet for all testing samples in figure \ref{fig:Reaction_Error_vs_Uncertainty}, where we also highlight an outlier corresponding to an out-of-distribution function sample in the testing data set (shown in blue). As one may expect, the prediction error for this testing sample is significantly higher than the other samples. We identify and visualize this worst-case-scenario example for which the UQDeepONet prediction provides the highest relative $\curlyL_2$ error, which is equal to $5.5\%$. We also report the marginal density of prediction error versus uncertainty estimates generated by the UQDeepONet, indicating that the UQDeepONet is able to provide calibrated and conservative uncertainty estimates. Figure \ref{fig:Reaction_max} presents a comparison between the ground truth solution and the UQDeepONet predictive mean and uncertainty. Although the prediction is still relatively accurate, we note that the form of the ground truth solution is quite different from most examples in the training and testing data-set; a fact that may justify why the model provides the maximum error for this case. We observe that the uncertainty obtained from the UQDeepONet model is much higher for this case, compared to the rest of samples shown in figure \ref{fig:Reaction_realizations}. This is an indication that the UQDeepONet method can not only provide accurate predictions when employed for a previously unseen example, but also allow one to identify out-of-distribution examples in the testing data-set.

\subsection{Reliable uncertainty estimates}\label{sec:BurgersExamples}

The Burgers' equation \cite{burgers1948mathematical} is well-known for modeling advective and dissipative behavior in fluids \cite{whitham2011linear}, nonlinear acoustics \cite{hamilton1998nonlinear}, and traffic flow \cite{lighthill1955kinematic}. It is studied as a prototype of turbulent flow and has the property of shock formation at the inviscid limit. The system of equations takes the form
\begin{equation}\label{equ:Burgers_equ}
\begin{aligned}
    \frac{\partial s}{\partial t} + s\frac{\partial s}{\partial x} &= \nu\frac{\partial^2 s}{\partial x^2}, \quad\quad\quad (x, t) \in [0,1]\times [0,1], \\
    s(x, 0) &= u(x),\quad\quad  x\in [0,1], \\
    s(0, t) &= s(1,t),\quad\quad  t\in [0,1], \\
    \frac{\partial s(0, t)}{\partial x} &= \frac{\partial s(1, t)}{\partial x},\quad\quad  t\in [0,1], \\
\end{aligned}
\end{equation}
where $\nu = 0.01$ a viscosity parameter. For the Burgers' equation, the input and out domains are $\curlyX$ and $\curlyY$, respectively, with $d_x = 1$ and $d_y = 2$. In this problem, we also map scalar functions, meaning $d_u = d_s = 1$, and the operator we are approximating is $\curlyG: C(\curlyX; \R) \to C(\curlyX; \R)$.

\begin{figure}
\centering
\includegraphics[width=0.9\textwidth]{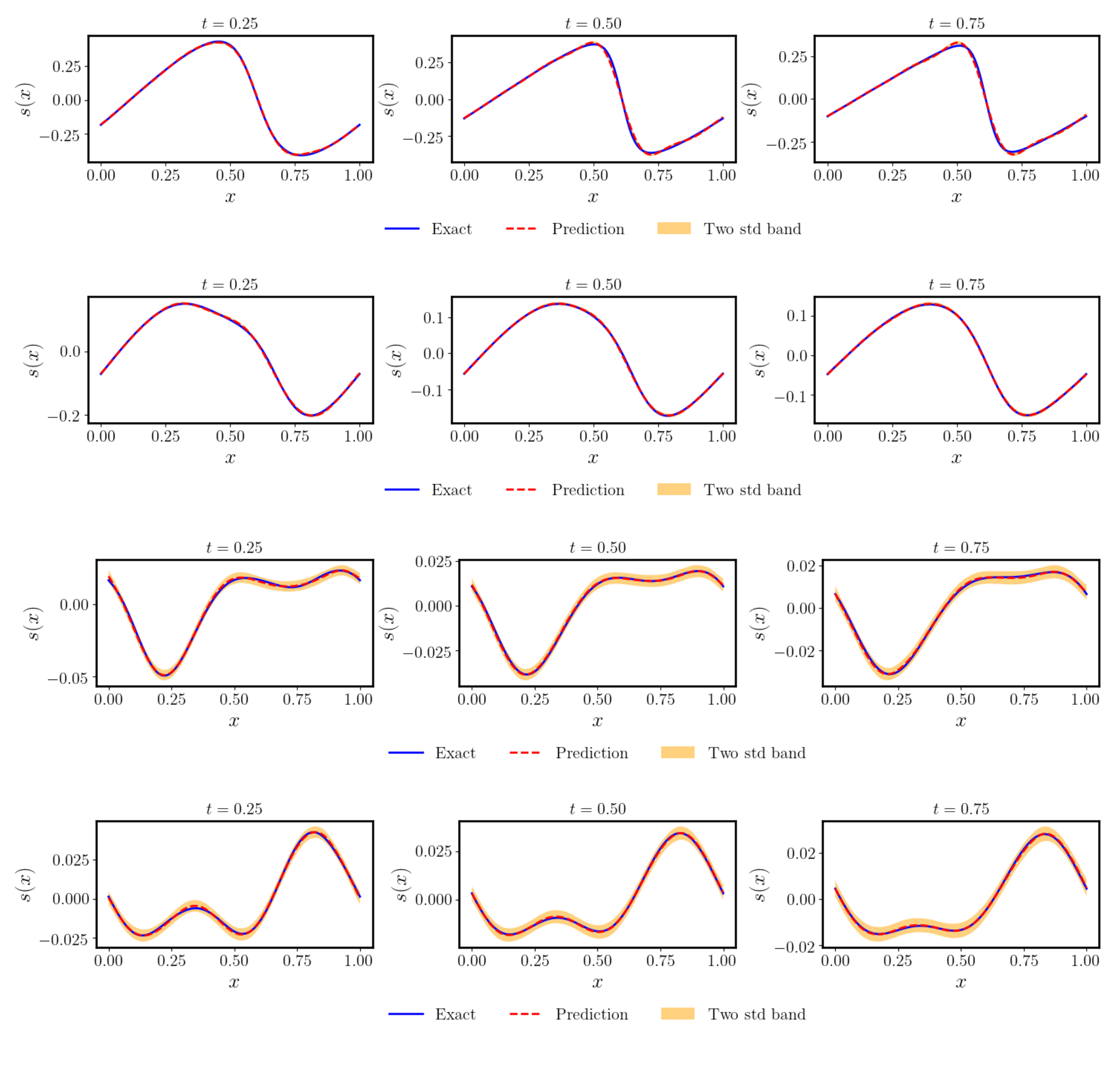}
\caption{{\em Burgers' transport dynamics:} UQDeepONet predictions for representative examples in the test data-set (top to bottom rows).  For each example, we present a comparison between the ground truth solution (blue), the predictive mean of the UQDeepONet (red dashed), and the predictive UQDeepONet uncertainty (yellow shade), for $t = 0.25$, $t = 0.50$ and $t = 0.75$.}
\label{fig:Burgers_realizations}
\end{figure}

The training and testing data generation process is summarized in the supplementary materials (see section \ref{sec:data_creation_supp}). Representative results obtained from a trained UQDeepONet are presented in figure \ref{fig:Burgers_realizations} where all testing errors are around $3\%$. For the top two samples, we observe a very accurate prediction and tight uncertainty bounds. For the bottom two samples, we get larger uncertainty in regions where the UQDeepONet predictive mean is not so accurate. These two examples demonstrate the ability of UQDeepONet to provide sensible predictions with reliable uncertainty estimates that reflect how confident the model is across different examples in the testing data-set.

\subsection{Handling noisy data}
\label{sec:SensorsExamples}

In our last example we turn our attention to a realistic climate modeling task where our goal is to learn a black-box operator that captures the functional relation between the air temperature and pressure fields on the surface of the Earth. This mapping takes the form
\begin{equation}
    T(x) \mapsto P(y),
\end{equation}
where $x,y \in [-90, 90] \times [0, 360]$ correspond to latitude and longitude coordinates charting the Earth's surface. In this case, the input and output function domains are the same, i.e. $\curlyX = \curlyY$, $d_x = d_y = 2$, and for the scalar temperature and pressure fields $d_u = d_s = 1$. Therefore the operator we aim to learn is $\curlyG: C(\curlyX; \R) \to C(\curlyX ; \R)$.

To enhance the performance of UQDeepONet for this benchmark, we exploit the periodic nature of the data and introduce a Harmonic Feature Expansion, as proposed by Lu {\it et. al.} \cite{lu2021comprehensive}
\begin{align}
    \xi(\bm{y}) = [\bm{y}, \sin(2\pi\bm{y}), \cos(2\pi\bm{y}), ..., \sin(2^H\pi\bm{y}), \cos(2^H\pi\bm{y})],
\end{align}
where $H$ is the order of the harmonic basis, which we set equal to $5$. Our experience indicates that Harmonic Feature Expansion significantly increases the accuracy of the UQDeepONet prediction for this benchmark by mitigating spectral bias \cite{wang2021eigenvector}, as the underlying input and output functions exhibit higher frequencies. 

\begin{figure}
\centering
\includegraphics[width=\textwidth]{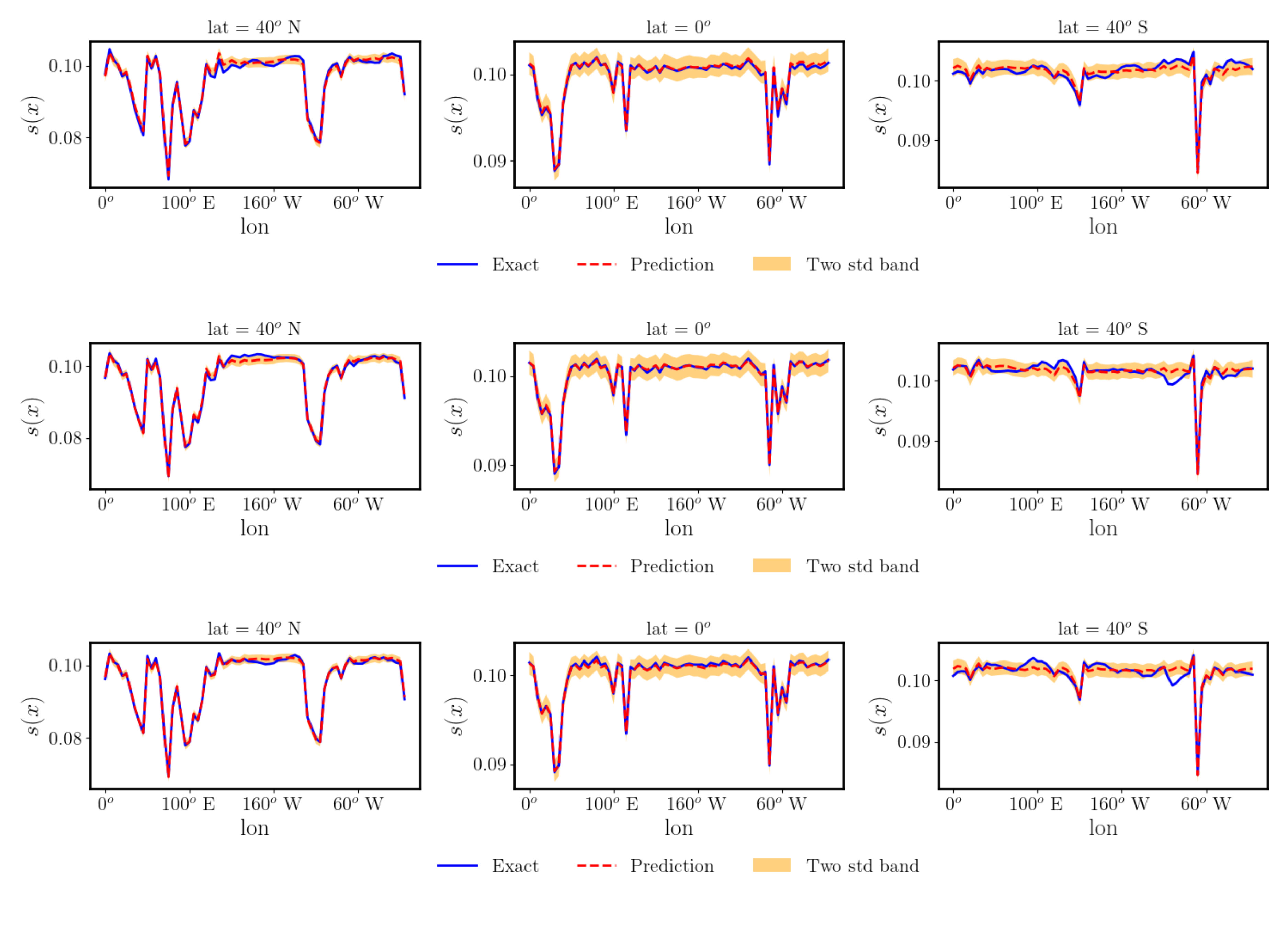}
\caption{{\em Climate modeling:} UQDeepONet predictions for representative examples in the test data-set (top to bottom rows).  For each example, we present a comparison between the reference surface air pressure (blue), the predictive mean of the UQDeepONet (red dashed), and the predictive UQDeepONet uncertainty (yellow shade), for latitude = $40^o N$, latitude = $0^o$ and latitude = $40^o S$.}
\label{fig:Weather_realizations}
\end{figure}

The construction of the training and testing data-sets is summarized in the supplementary material (see section \ref{sec:data_creation_supp}). Representative predictions of a trained UQDeepONet are presented in figure \ref{fig:Weather_realizations}. We observe that the model is quite accurate and provides tight confidence bounds in regions where the target output functions are relatively smooth and well-behaved, while the predictive uncertainty is larger in regions where the output functions are more oscillatory.

\section{Discussion}\label{sec:discussion}
In this work, we propose a novel uncertainty quantification approach for operator learning, called UQDeepONet. The proposed method is tested for two synthetic examples of learning the operator of a non-linear partial differential equation, one example of learning the anti-derivative operator and  one realistic climate modeling example where an unknown operator between the surface air temperature and surface air pressure is approximated. With these examples, we experimentally demonstrate that the proposed model provides accurate predictions for most random function pairs in a never-seen data-set as well as reasonable uncertainty estimates for outliers. Moreover, taking advantage of the computational efficiency of randomized prior networks, we can train ensembles consisting of hundreds of DeepONet in a few hours on a single GPU. Taken together, these developments provide advanced computational infrastructure that enables surrogate modeling and uncertainty quantification for infinite-dimensional parametric problems, as the inputs to our model are, in principle, continuous functions.

Beyond the results presented in this paper, there are few directions that need further investigation. First, while the current UQDeepONet method is mostly based on the original DeepONet architecture of Lu {\it et al.} \cite{lu2021learning}, it would be interesting to extend it to include the modifications recently proposed in \cite{lu2021comprehensive,wang2021improved} that have been empirically demonstrated to yield better performance. Second, randomized priors are not a concept is only applicable to the DeepONet model, but could be also enable scalable uncertainty quantification for other operator learning frameworks, such as the Fourier Neural Operator \cite{li2020fourier} or the attention-based architecture LOCA \cite{kissas2022learning}. Third, several advanced tools for efficient optimization and stabilization of the training dynamics, such as the Neural Tangent Kernel approach \cite{jacot2018neural, wang2021improved} and hierarchical incremental gradient descent \cite{su2018uncertainty}, could be considered for achieving better predictive accuracy and further accelerate the training process of multiple neural networks. One current limitation that warrants attention in future work is the limited ability of DeepONet and other operator learning techniques to receive input functions whose domain is very high-dimensional. To this end, dimension reduction or low-rank approximation techniques can be employed to reduce computational complexity. Finally, while uncertainty quantification is predominately used for providing prediction error bars and confidence intervals, it also plays a prominent role in decision making and the judicious acquisition of new data. To this end, the question of how to acquire informative data for operator learning tasks is currently under-explored and calls for further investigation in the future.

\section*{Acknowledgements}
The authors  would like to acknowledge support from the US Department of Energy under the Advanced Scientific Computing Research program (grant DE-SC0019116), the US Air Force (grant AFOSR FA9550-20-1-0060), and US Department of Energy/Advanced Research Projects Agency (grant DE-AR0001201). We also thank the developers of the software that enabled our research, including JAX \cite{jax2018github}, Matplotlib \cite{hunter2007matplotlib} and NumPy \cite{harris2020array}.

\bibliographystyle{unsrt}
\bibliography{main.bib}

\newpage

\appendix

\section{Parallel implementation and computational cost}\label{sec:parallel_cost_supp} 

We demonstrate the performance of our implementation for the anti-derivative benchmark by presenting the wall-clock time needed for the training of a UQDeepONet with different ensemble sizes in table \ref{tab:time_cost_supp}. We observe that the method is computationally efficient as the cost exhibits a near-linear scaling with the number networks in the ensemble, e.g. the wall-clock time for training 64 DeepONet on a single GPU is just 4 times larger than training a single DeepONet. Moreover, the parallel training of a UQDeepONet ensemble with 512 networks finishes within half an hour, which indicates the efficiency of the method. The computational cost mainly consists of two parts: (1) the cost for training the neural networks in the UQDeepONet ensemble, (2) the cost of fetching different mini-batches of training data at each gradient descent iteration. We notice that for small ensemble sizes, the cost of training neural networks is roughly in the same order of the cost needed to fetch data batches, indicating that a small number of networks is not sufficient to efficiently utilize the full capacity of a GPU. To this end, better hardware utilization is observed as the ensemble size is increased, indicating better parallel performance and the observed linear scaling.

\begin{table*}[!ht]
    \centering
    \caption{Training time of a UQDeepONet for different ensemble sizes.}\label{tab:time_cost_supp}
\resizebox{\textwidth}{!}{%
    \begin{tabular}{l|cccccccccc}
      \toprule 
        Ensemble size & \bfseries 1 & \bfseries 2 & \bfseries 4 & \bfseries 8 & \bfseries 16 & \bfseries 32 & \bfseries 64 & \bfseries 128 & \bfseries 256 & \bfseries 512\\ 
      \midrule 
      Training time (sec) & 48.66 & 56.52 & 53.45 & 53.09 & 68.46 & 106.11 & 172.53 & 315.74 & 597.74 & 1160.53 \\
      \bottomrule 
    \end{tabular}
}
\end{table*}

\section{Effect of the scaling parameter $\beta$}\label{sec:anti_studybeta_supp}
\begin{figure}[H]
\centering
\includegraphics[width=\textwidth]{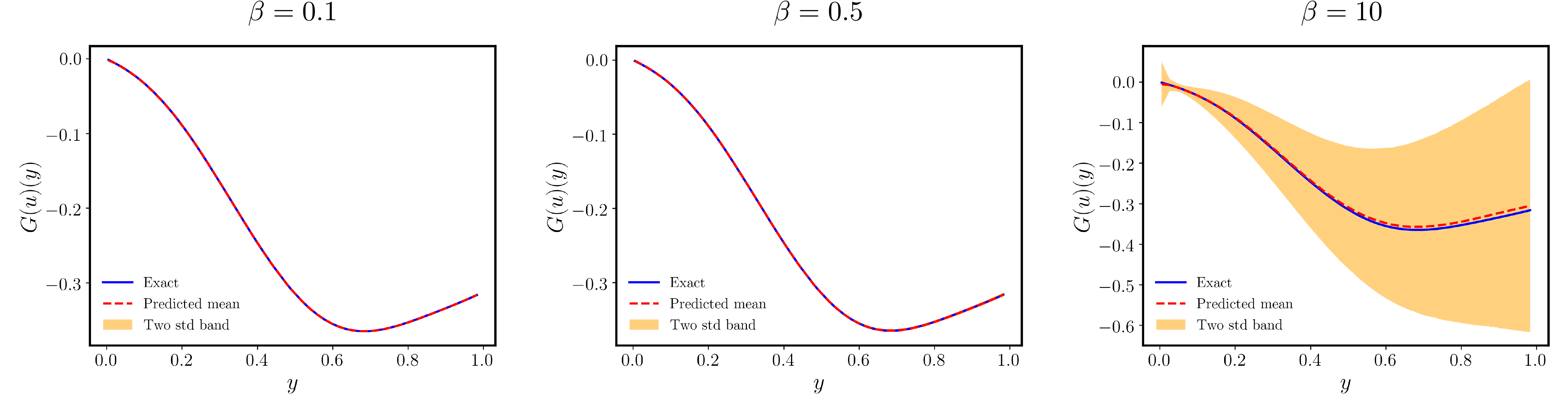}
\caption{{\em Effect of the scaling parameter $\beta$:} Prediction of a trained UQDeepONet for a randomly chosen example in the test data-set of the anti-derivative benchmark. From left to right: $\beta = 0.1$, $\beta =0.5$ and $\beta =10$. The red dashed lines represent the predicted mean, the blue solid lines represent the reference solution, and the yellow shaded areas represent two standard deviations of the UQDeepONet predictions.}
\label{fig:Anti_Beta}
\end{figure}

In this section, we study how the different values of the $\beta$ hyper-parameter in the randomized prior networks affects the model output (see equations \ref{equ:UQDeepONet_branch_trunk_1} and \ref{equ:UQDeepONet_branch_trunk_2}).  While the default $\beta$ value used in the literature is $1$ \cite{osband2018randomized, osband2021evaluating, ciosek2019conservative}, here we examine the performance of different setting, such as $0.1, 0.5$ and $10$, to test the sensitivity of the results this choice. Figure \ref{fig:Anti_Beta}, shows the prediction of UQDeepONet on a randomly chosen testing input/output function pair. We observe that the results are not sensitive to $\beta$ when it takes values in $\mathcal{O}(1)$, while an excessively large value of $\beta$ will inevitably lead to increasingly conservative uncertainty estimates. This is expected as the contribution of the fixed prior networks in equations \ref{equ:UQDeepONet_branch_trunk_1} and \ref{equ:UQDeepONet_branch_trunk_2} will exhibit increasingly high variance,  casting a great difficulty to the trainable network to fit data. In practice, for the purpose of obtaining reliable uncertainty estimates, the fixed prior network is assumed to have the same scale as the trainable network after being properly initialized. Without a loss of generality, we adopt the default setting used in the literature and set $\beta = 1.0$ throughout all experiments in this manuscript. 

\section{Data generation}\label{sec:data_creation_supp}
In this section, we provide details on how we generate the training and testing data for all numerical experiments considered in this work.

\subsection{Reaction-diffusion dynamics}
A training data-set is constructed by generating random source term functions $u(x)$ from a Gaussian processes prior \cite{rasmussen2003gaussian} with an exponential quadratic kernel using a length-scale parameter of $l=0.2$. Each realization is then recorded on an equi-spaced grid of $m=500$ sensor locations. We numerically solve the system of equations \eqref{equ:Reaction_equ} using second order finite differences to obtain discrete measurements on a $100 \times 100$ grid for training. As such, we generate $N = 1,000$ input/output function pairs for training the UQDeepONet model, and another $1,000$ pairs for testing its performance.

\subsection{Burgers' transport dynamics}
Training and testing data-sets are generated by sampling random initial condition functions $u(x)$ from a Gaussian Random Field $u \sim \mathcal{N}(0, 25^2(-\Delta + 25I)^{-2})$ on a $100 \times 100$ grid, and using them as input to solve the Burgers' equation using a spectral numerical solver (Chebfun package in Matlab \cite{driscoll2014chebfun}) to obtain ground truth estimates for the target solution $s(x,t)$. As such, we construct a training data-set consisting of $N=1,000$ input/output function pairs, and a testing data-set consisting of another $N=1,000$ pairs.

\subsection{Climate modeling}
Training and testing data-sets are obtained from the Physical Sciences Laboratory database \cite{kalnay1996ncep} \footnote{\url{https://psl.noaa.gov/data/gridded/data.ncep.reanalysis.html}} which contains daily measurements of surface air temperature and pressure from years 2000 to 2010. Our training data-set is constructed by taking measurements from 2000 to 2005 to obtain $N= 1,825$ input/output function pairs evaluated on a $2.5$ degree latitude and $2.5$ degrees longitude ($144 \times 72$) grid, which we then sub-sample to a $72 \times 72$ grid. Similarly, our testing data-set comprises daily measurements from 2005 to 2010 which are pre-processed in an equivalent manner to obtain $N=1,825$ input/output function pairs on a $72 \times 72$ grid.

\section{Out-of-distribution detection}
\label{sec:appendix}
Similar to the reaction-diffusion equation example in section \ref{sec:ReactionExamples}, here we also report  out-of distribution detection results for the Burgers' transport and climate modeling benchmarks.

\subsection{Burgers' transport dynamics}\label{sec:Burgers_supp}
\begin{figure}
\centering
\includegraphics[width=0.9\textwidth]{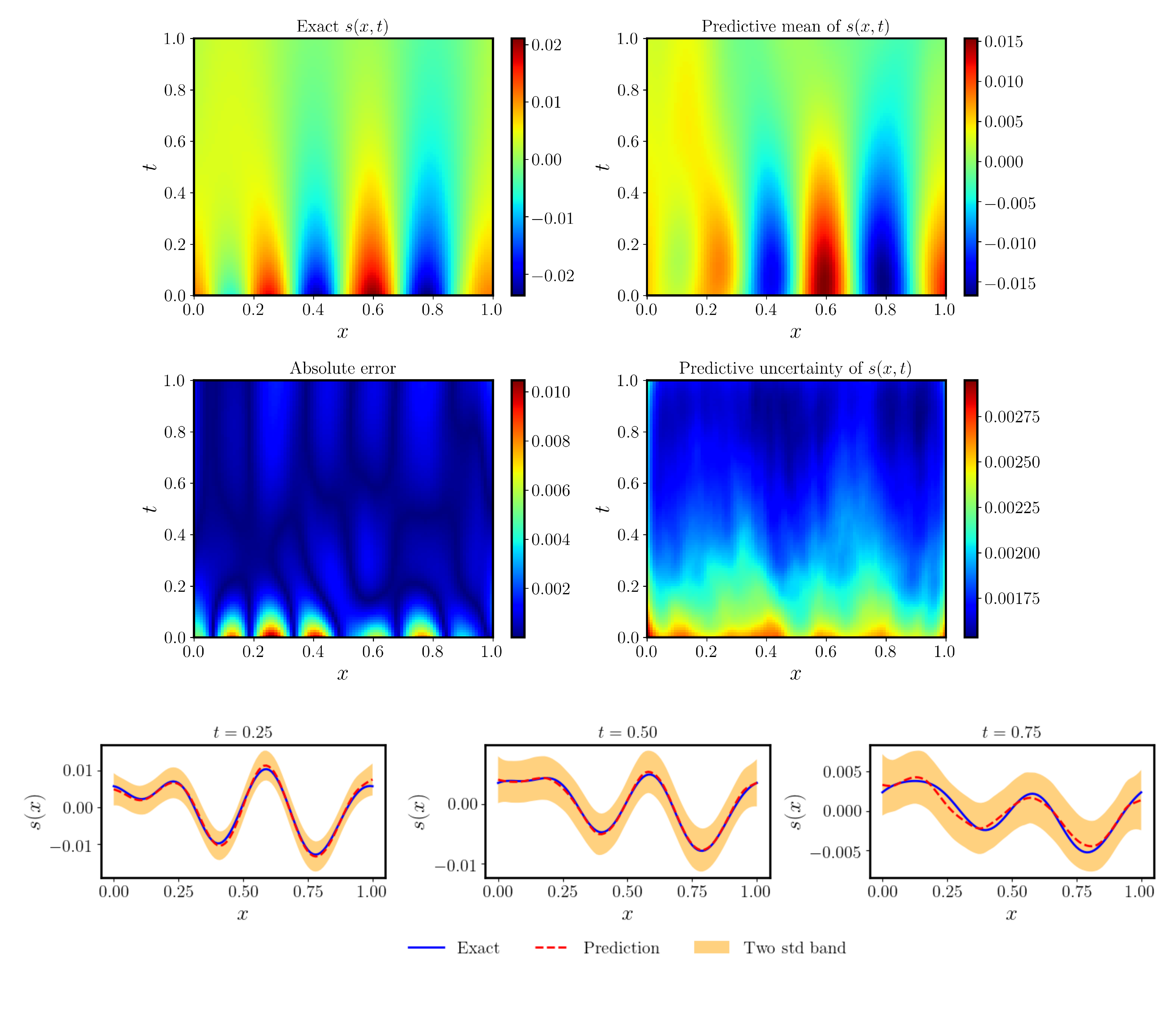}
\caption{{\em Burgers' transport dynamics:} Worst case scenario prediction. Top panel: Ground truth solution and UQDeepONet predictive mean. Middle panel: Absolute error between the ground truth solution and the UQDeepONet predictive mean, as well as the UQDeepONet predictive uncertainty. Bottom panel: Ground truth solution (blue), UQDeepONet predictive mean (red dashed) and uncertainty (yellow shade) at $t = 0.25$, $t = 0.50$ and $t = 0.75$.}
\label{fig:Burgers_max}
\end{figure}

We identify the function sample in the testing data-set for which the UQDeepONet prediction provides the highest relative error, equal to $21.6\%$. For this example, Figure \ref{fig:Burgers_max} presents a comparison between the ground truth solution and the UQDeepONet predictive mean, as well as the absolute point-wise error and the corresponding UQDeepONet predictive uncertainty. We observe that the target output function in this case not only exhibits very small magnitude, but also exhibits different frequencies than most examples in the training and testing data-sets. This justifies the poor predictive mean predictions we see in this case. However, the lack of accuracy is clearly reflected in the high predictive uncertainty reported by the UQDeepONet model, whose distribution seems to correlate well with the absolute point-wise error. In realistic scenarios where the ground truth solution may be unknown, we argue that one cannot simply rely on using deterministic DeepONet predictions as those provide no indication of whether the model is confident in what it predicts or not. This highlights the crucial aspect of uncertainty quantification provided by UQDeepONet, which makes them suitable to adopt for risk-sensitive applications.

\subsection{Climate modeling}\label{sec:Climate_supp}
\begin{figure}
\centering
\includegraphics[width=\textwidth]{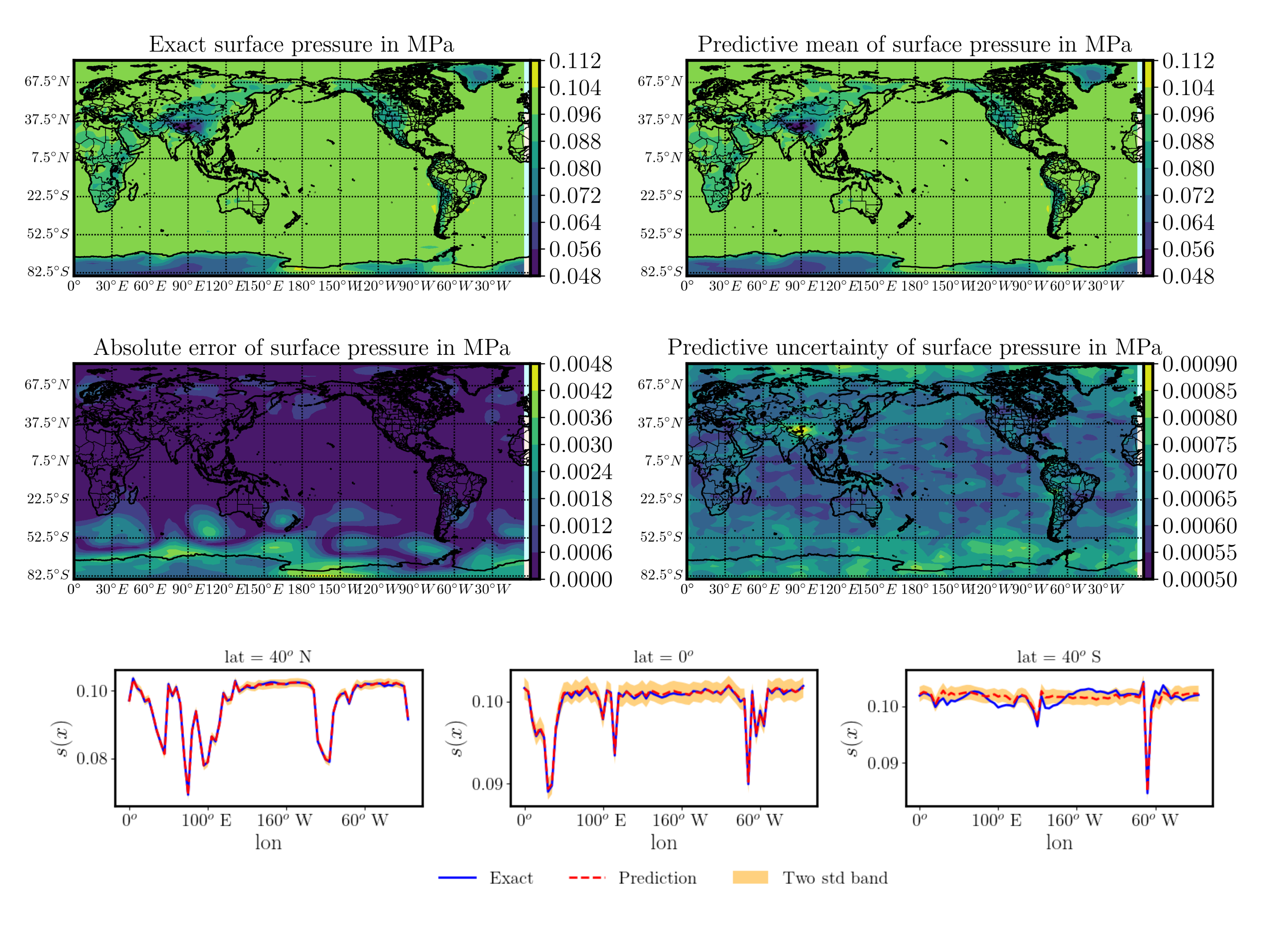}
\caption{{\em  Climate modeling:} Worst case scenario prediction. Top panel: Reference surface air pressure and UQDeepONet predictive mean. Middle panel: Absolute error between the reference surface air pressure and the UQDeepONet predictive mean, as well as the UQDeepONet predictive uncertainty. Bottom panel: Reference surface air pressure (blue), UQDeepONet predictive mean (red dashed) and uncertainty (yellow shade) at latitude = $40^o N$, latitude = $0^o$ and latitude = $40^o S$.}
\label{fig:Weather_max}
\end{figure}

We identify the function sample in the testing data-set that the UQDeepONet model provides the highest relative error, equal to $13.1\%$. We present a comparison between the model prediction and the baseline output function, as well as the absolute error and the uncertainty estimate in figure \ref{fig:Weather_max}. Moreover, we observe that the model provided realistic uncertainty estimates nearly everywhere that there is a discrepancy between the predicted mean and the baseline measurements. In this example, both the measurements of the input and the output functions are noisy, which motivates the use of UQDeepONet in real world applications where learning a black-box operator is needed. 

\end{document}